\documentclass{article}

\pdfoutput=1

\usepackage[preprint]{neurips_2023}

\usepackage[utf8]{inputenc} 
\usepackage[T1]{fontenc}    
\usepackage{hyperref}       
\usepackage{url}            
\usepackage{booktabs}       
\usepackage{amsfonts}       
\usepackage{nicefrac}       
\usepackage{microtype}      
\usepackage{xcolor}         
\usepackage{enumitem}       
\usepackage{amsthm,amsmath,amssymb} 
\usepackage{mathrsfs}
\usepackage{graphicx}       
\usepackage{subcaption}     
\usepackage{epstopdf}
\usepackage{multirow}       
\usepackage{tabularx}       
\usepackage{algorithm}      
\usepackage{algpseudocode}  
\newcolumntype{Z}{>{\centering\let\newline\\\arraybackslash\hspace{0pt}}X} 
\setcitestyle{numbers,square,comma}
\usepackage{hyperref} 
\usepackage{epstopdf} 

\title{Self-discipline on multiple channels}

\author{
    Jiutian Zhao$^{1,2}$ \\ \texttt{zhaojiutian@whut.edu.cn} \And
    Liang Luo$^{1,2}$ \\ \texttt{luoliang@whut.edu.cn} \And
    Hao Wang$^{1,2}$ \\ \texttt{hao\_wang@whut.edu.cn} \and \\
  $^1$Key Laboratory of High-Performance Ship Technology (Wuhan University of Technology), \\Ministry of Education, Wuhan 430063, China\\
  $^2$School of Naval Architecture, Ocean and Energy Power Engineering, \\Wuhan University of Technology, Wuhan 430063, China \\
}

\begin{document}

\maketitle

\begin{abstract}
    Self-distillation relies on its own information to improve the generalization ability of the model and has a bright future. Existing self-distillation methods either require additional models, model modification, or batch size expansion for training, which increases the difficulty of use, memory consumption, and computational cost. This paper developed Self-discipline on multiple channels(SMC), which combines consistency regularization with self-distillation using the concept of multiple channels. Conceptually, SMC consists of two steps: 1) each channel data is simultaneously passed through the model to obtain its corresponding soft label, and 2) the soft label saved in the previous step is read together with the soft label obtained from the current channel data through the model to calculate the loss function. SMC uses consistent regularization and self-distillation to improve the generalization ability of the model and the robustness of the model to noisy labels. We named the SMC containing only two channels as SMC-2. Comparative experimental results on both datasets show that SMC-2 outperforms Label Smoothing Regularizaion and Self-distillation From The Last Mini-batch on all models, and outperforms the state-of-the-art Sharpness-Aware Minimization method on 83\% of the models.Compatibility of SMC-2 and data augmentation experimental results show that using both SMC-2 and data augmentation improves the generalization ability of the model between 0.28\% and 1.80\% compared to using only data augmentation. Ultimately, the results of the label noise interference experiments show that SMC-2 curbs the tendency that the model's generalization ability decreases in the late training period due to the interference of label noise. The code is available at \href{https://github.com/JiuTiannn/SMC-Self-discipline-on-multiple-channels}{https://github.com/JiuTiannn/SMC-Self-discipline-on-multiple-channels}.
\end{abstract}

\section{Introduction}

In 2015 Hinton et al. \cite{hinton2015distilling} addressed the problems in deep learning by proposing knowledge distillation (KD) using the migration of complex deep network models to shallow, small network models. After several years of development, there are several ways of distillation learning: offline distillation \cite{hinton2015distilling,ahn2019variational} where the student model learns based on a pre-trained teacher model with fixed parameters; online distillation \cite{zhang2018deep,zhu2018knowledge,song2018collaborative} where both teacher and student model parameters are updated; multi-model distillation \cite{you2017learning,park2019feed} where multiple models participate in distillation; data-free distillation \cite{lopes2017data,fang2019data,nayak2019zero} where distillation is performed without using any known dataset; privileged distillation \cite{wang2018kdgan,tang2019retaining} where the teacher has some information that is not accessible to the student privileged distillation where the teacher has some special information constraints that the students do not have access to; self-distillation \cite{du2022sharpness,shen2022self,zhang2019your,phuong2019distillation,lee2020self,kim2020self} learning where students distill on their own information alone without relying on other models.

In self-distillation, BYOT \cite{zhang2019your} and DBA \cite{phuong2019distillation} need to modify the network structure, which greatly increases the difficulty of using the method; SKBG \cite{kim2020self} and SAF \cite{du2022sharpness} use the information from their own previous epochs to guide the current model, but the information may be somewhat "outdated" \cite{shen2022self}, and such methods are implemented by taking a snapshot of the model or introducing the same model as their own, which undoubtedly increases the memory usage. Recently, DLB \cite{shen2022self} has solved the above problem, but its batch size has increased to twice of the original size, and it computes two loss functions with different batch sizes, which reduces the accuracy to some extent. Consistent regularization, which is widely used in semi-supervised learning, requires the introduction of extra models or model snapshots \cite{laine2016temporal}, which increases memory consumption.

Therefore, we propose the self-discipline on multiple channels (SMC) method to address the problems of self-distillation and semi-supervised learning.

Our contributions are as follows:

\begin{itemize}[leftmargin = 20 pt]

     \item We propose the SMC method to improve the generalization effect of neural networks. SMC improves the generalization ability of the model better than LSR \cite{szegedy2016rethinking}, DLB and SAM \cite{foret2020sharpness}.

    \item The SMC approach proposes the idea of multiple data channels that can enhance generalization using multiple principles at the same time(e.g. consistency regularization with self-distillation is used in this paper).

    \item We verified the effectiveness of SMC over image classification tasks by using AlexNet \cite{krizhevsky2014one}, VGG19-BN \cite{simonyan2014very}, ResNet-56, ResNet-110 \cite{he2016deep}, WRN28-10 \cite{zagoruyko2016wide}, PreResNet-110 \cite{he2016identity}, DenseNet-100-12 \cite{huang2017densely} in neural network models.

    \item We experimentally demonstrate the robustness of SMC against label noise, i.e., the correct rate does not increase and then decrease after using SMC and the generalization ability is improved.
\end{itemize}

\section{Preliminaries}

\subsection{Notations}

\begin{itemize}[leftmargin = 20 pt]

\item $ \mathfrak{D}=\{(x_i,y_i )\}_{i=1}^N $: A $ K $-class labeled dataset.N is the total number of training instances.

\item $ \mathcal{B}_t= \{x_i^t,y_i^t \}_{i=1}^n $: A training sample at the $ t $-th iteration in a training epoch, each sample $ \{x_i,y_i\} $ obeys the distribution $ \mathfrak{D} $. $ n $ is the batch size.

\item $ \mathcal{M}(x;\omega) $: A model of a neural network, where $ \omega\in\mathbb{R}^k $ are trainable parameters. $ k $ is the dimensionality of the parameters.

\item $ \mathcal{L} $: A loss function.

\item $ \mathcal{M}_k(x;\omega) $: The input $ x $ is the probability of the class $ k\in K $.

\item $ \omega_t $: Parameters of model $ \mathcal{M} $ at step $ t $.

\item $ \mathcal{C}_{A} $: All data in channel A.

\item $ \{p_i^{ \tau  , \theta }\}_{x_A} $: The soft labels obtained after the training samples $ x_A \in \mathcal{C}_{A} $ in channel A are input to $ \mathcal{M}(x;\omega_\theta ) $, where $ \tau $ is the distillation temperature.

\item $ \phi $: A data augmentation method.

\item $ \mathfrak{s} $: The total number of mini-batches trained at the current time.

\item $ \mathfrak{S} $: The number of mini-batches to be trained in the whole training.

\item $ \lambda $: The weight ratio between $ \mathcal{L}_{CE} $ and $ \mathcal{L}_{KL} $.

\item $ \alpha $: The maximum value of the weight of $ \mathcal{L}_{KL} $
\end{itemize}

\subsection{Related Works}


\paragraph{Self-Distillation from the Last Mini-Batch (DLB).}DLB \cite{shen2022self} is based on SKBG and reduces the interval between the preceding and following moments to mini-batch, which saves a lot of memory and improves the performance. Specifically, the new data $ [x_{t-1},x_t] $ is formed by reading the batch data $ x_{t-1} $ saved in the previous step together with the batch data $ x_t $ in the current step, and then calculating the cross-entropy loss $ \mathcal{L}_{CE} $ of the hard tags of $ [x_{t-1},x_t] $ and the Kullback-Leibler divergence loss $ \mathcal{L}_{KL} $ of the soft and hard tags of $ x_{t-1} $. However, this method will enlarge the batch size, increase the memory occupation, and calculate $ \mathcal{L}_{CE} $ and $ \mathcal{L}_{KL} $ are not the same batch size, we think it will lose accuracy.

\paragraph{Consistency Regularization.}The core idea of consistency regularization, which is widely used in semi-supervised learning \cite{laine2016temporal,bachman2014learning,luo2018smooth,tarvainen2017mean}, is that the prediction result should be constant for a network's input even if it suffers from small disturbances. TEF \cite{laine2016temporal} is the use of constructing additional $ \pi $ models or recording all soft labels of the dataset $ \mathfrak{D} $ to achieve consistency regularization, which greatly increases the difficulty of use or memory usage.

\paragraph{Sharpness-Aware Minimization (SAM).}The traditional training method for neural networks can converge to a good local optimum, but the vicinity of that point is often a steep valley and the generalisation performance of the model decreases when there is a small perturbation in the weights, so SAM \cite{foret2020sharpness} tends to find a locally smoothed minimum point via Equation \ref{equ:SAM}.
\begin{equation}
    {\underset{\omega}{\mathit{\min}}{\mathcal{L}_{SAM}(\omega) + \mu \parallel \omega \parallel_{2}^{2}}}~~where~~\mathcal{L}_{SAM}(\omega) \triangleq ~{\underset{\parallel \epsilon \parallel_{p} \leq \mathit{~\rho}}{\mathit{\max}}{\mathcal{L}_{SAM}\left( {\omega + \epsilon} \right)}}
\label{equ:SAM}    
\end{equation}
where $ p \in \left\lbrack {1,\infty} \right\rbrack $ and $ \rho \geq 0 $.

\section{SMC Methodology}

\subsection{Formulation}

\begin{figure}[htbp]
    \centering
    \includegraphics[height = 8cm, width = 12cm]{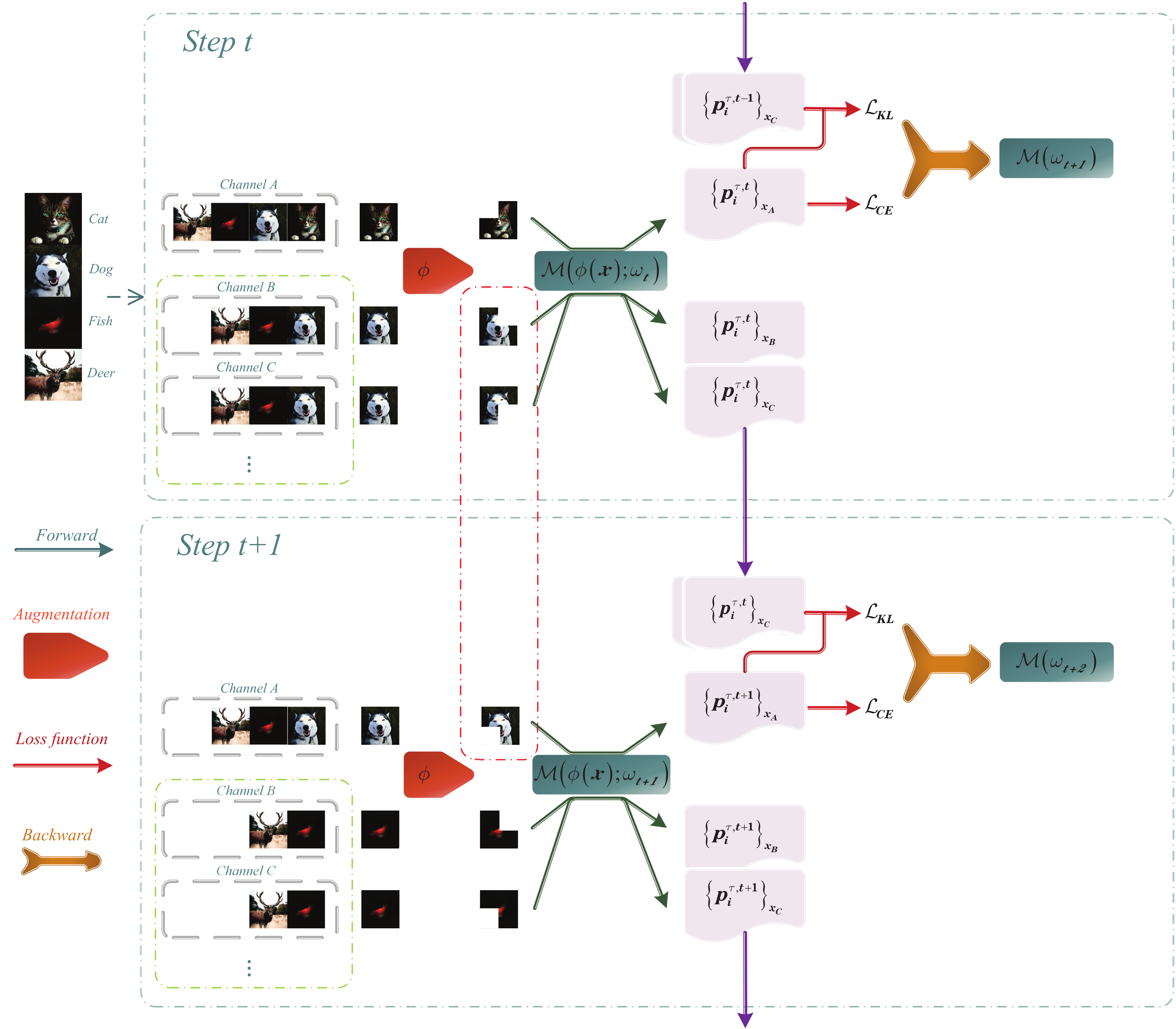}
    \caption{The architecture of SMC. Data augmentation using Cutout as an example.}
    \label{fig:SMCA}
  \end{figure}

As shown in Figure \ref{fig:SMCA},we propose the concept of multiple data channels to achieve consistency regularization and self-distillation. The data of the whole training process is represented as $ \left\{ {\left\{ {B_{1}{,B}_{2},B_{3},B_{4},\cdots,B_{N}} \right\},\mathfrak{D}_{N},\cdots,\mathfrak{D}_{N}} \right\} $. All data in channel A:
\begin{equation}
    \mathcal{C}_{A} = \left\{ {\left\{ {B_{1}{,B}_{2},B_{3},B_{4},\cdots,B_{N}} \right\},\mathfrak{D}_{N},\cdots,\mathfrak{D}_{N}} \right\}
\label{equ:CA}    
\end{equation}
All data in Channel A, Channel B, etc:
\begin{equation} 
    \left\{ \begin{matrix}
        {\mathcal{C}_{B} = \left\{ {\left\{ {\mathcal{B}_{2},\mathcal{B}_{3},\mathcal{B}_{4},\cdots,\mathcal{B}_{N}} \right\},\mathfrak{D}_{N},\cdots,\mathfrak{D}_{N}} \right\}} \\
        {\mathcal{C}_{C} = \left\{ {\left\{ {\mathcal{B}_{2},\mathcal{B}_{3},\mathcal{B}_{4},\cdots,\mathcal{B}_{N}} \right\},\mathfrak{D}_{N},\cdots,\mathfrak{D}_{N}} \right\}} \\
         \vdots \\
        \end{matrix} \right.
\label{equ:CBC}    
\end{equation}
where $ \left\{ \mathcal{B}_{i} \right\}_{i = 1}^{N} = ~\mathfrak{D}_{N} $.

We denote the trained neural network model by $ \mathcal{M}\left( \phi(x);\omega_{t} \right) $. $ \phi(x) $ is the result of the data augmentation on the data $ x $.$ \omega_{t} $ is the parameter of the neural network model at step $ t $. Note that the same data augmentation method is used for the data in multiple channels, and there are random events in the augmentation method, so the augmented data results are likely to be different.

\subsection{Method Architecture}

SMC does not require additional models or modifications to the model when used as in TEF \cite{laine2016temporal}. SMC takes advantage of the random events present in data augmentation techniques to produce different augmentation effects $ \phi_{A}(x) , \phi_{B}(x) , \phi_{C}(x)$, etc. for the same data $ x~\epsilon~\mathfrak{D} $. The soft label corresponding to each channel data is obtained by Equation \ref{equ:MAB}.
\begin{equation}
    \left\{ \begin{matrix}
    {\left\{ p_{i}^{\tau,t} \right\}_{x_{A}} = ~M\left( {\phi_{A}(x);\omega_{t}} \right)~} \\
    {\left\{ p_{i}^{\tau,t - 1} \right\}_{x_{B}} = ~M\left( {\phi_{B}(x);\omega_{t - 1}} \right)} \\
    {\left\{ p_{i}^{\tau,t - 1} \right\}_{x_{C}} = ~M\left( {\phi_{C}(x);\omega_{t - 1}} \right)} \\
    \vdots \\
    \end{matrix} \right.
\label{equ:MAB}    
\end{equation}
The Kullback-Leibler divergence loss of $ \mathcal{L}_{KL} $ of $ \left\{ p_{i}^{\tau,t} \right\}_{x_{A}} $ , $ \left\{ p_{i}^{\tau,t - 1} \right\}_{x_{B}} $ , $ \left\{ p_{i}^{\tau,t - 1} \right\}_{x_{C}} $, etc. is calculated by Equation \ref{equ:LKL} for consistency regularization and self-distillation.
\begin{equation}
    \left\{ \begin{matrix}
        \begin{matrix}
        {\mathcal{L}_{KL}^{1} = \frac{\lambda}{\left| \mathcal{B} \right|}{\sum\limits_{i = 1}^{n}{KL\left( {\left\{ p_{i}^{\tau,t} \right\}_{x_{A}},\left\{ p_{i}^{\tau,t - 1} \right\}_{x_{B}}} \right)}}} \\
        {\mathcal{L}_{KL}^{2} = \frac{\lambda}{\left| \mathcal{B} \right|}{\sum\limits_{i = 1}^{n}{KL\left( {\left\{ p_{i}^{\tau,t} \right\}_{x_{A}},\left\{ p_{i}^{\tau,t - 1} \right\}_{x_{C}}} \right)}}} \\
        \end{matrix} \\
         \vdots \\
        \end{matrix} \right.
\label{equ:LKL}    
\end{equation}
where $ \tau $ is the distillation temperature, and the label is flatter when $ \tau $ is higher. Previous works \cite{szegedy2016rethinking,yuan2020revisiting} suggested that the introduction of $ \mathcal{L}_{KL} $ in knowledge distillation produces effects similar to label smoothing, but recent works \cite{szegedy2016rethinking} suggested that it has some effect of reducing the loss of sharpness.

As with self-distillation, in order to learn from hard label $ y_{A} $, the Cross-Entropy loss $ \mathcal{L}_{CE} $ of $ y_{A} $ and $ \left\{ p_{i}^{\tau,t} \right\}_{x_{A}} $ are calculated by Equation \ref{equ:LCE}
\begin{equation}
\mathcal{L}_{CE} = \frac{1}{n}{\sum\limits_{i = 1}^{n}{H\left( {y_{A},\left\{ p_{i}^{\tau,t} \right\}_{x_{A}}} \right)}}
\label{equ:LCE} 
\end{equation}
where $ \left\{ p_{i}^{\tau,t} \right\}_{x_{A}} = \left( q_{j}(1),\cdots,q_{j}(K) \right) $ is the soft label generated by sample $ x_{A} $ through model $ \mathcal{M}\left( {x;\omega_{t}} \right) $

\begin{figure}[htbp]
    \centering
    \includegraphics[height = 2cm, width = 3.5cm]{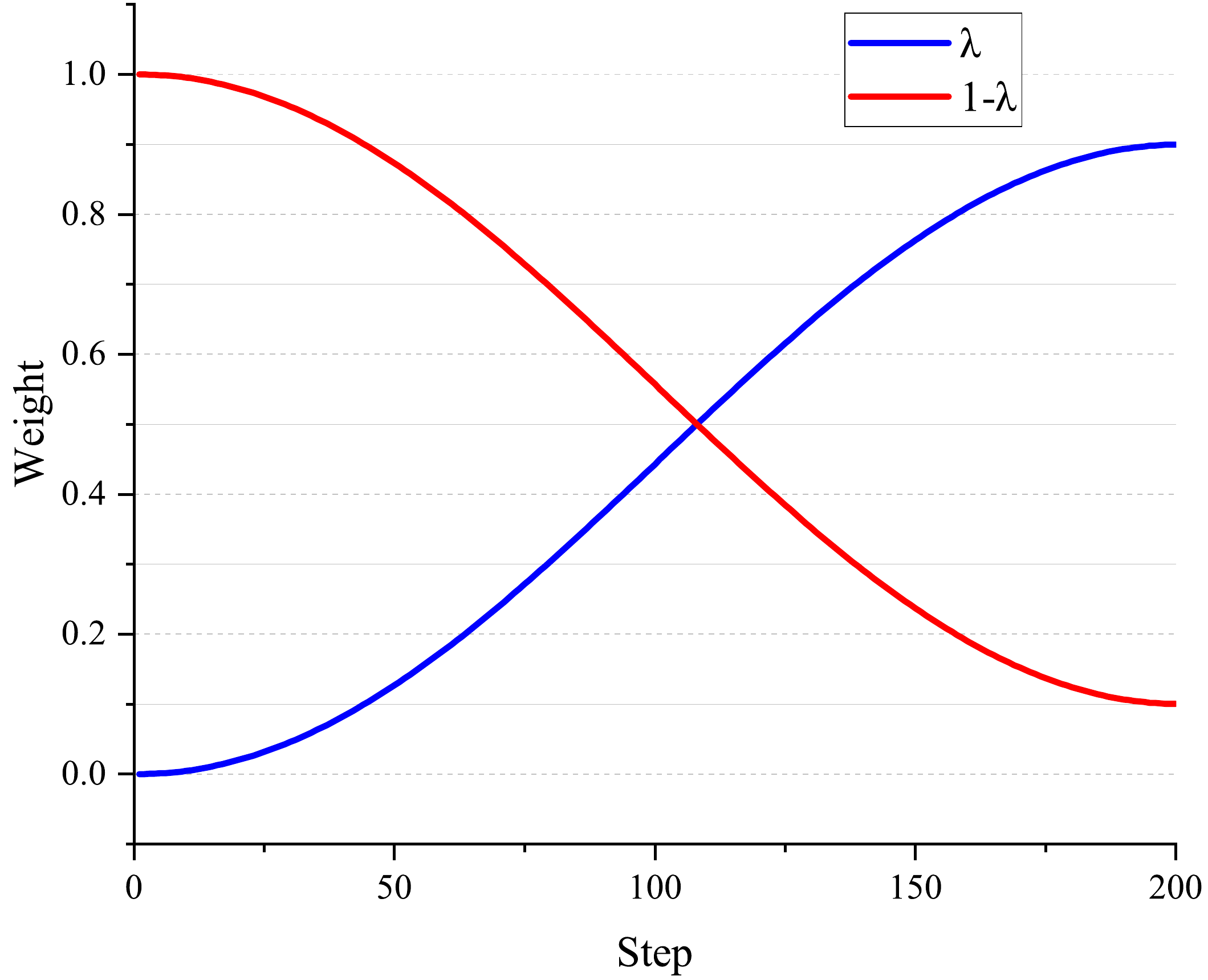}
    \caption{When $ \alpha = 0.9 $, the change in the weights of $ \mathcal{L}_{KL} $ and $ \mathcal{L}_{CE} $. In the figure, $ \alpha $ is the weight of $ \mathcal{L}_{KL} $ and $ 1 - \alpha $ is the weight of $ \mathcal{L}_{CE} $}
    \label{fig:lam}
\end{figure}

To calculate the direction of the overall gradient descent, we adjust the guiding effect of $ \mathcal{L}_{CE} $ and $ \mathcal{L}_{KL} $ on the overall gradient by the weights $ \lambda $ in Equation \ref{equ:L}
\begin{equation}
\mathcal{L} = \left( {1 - \lambda} \right)\mathcal{L}_{CE} + ~~\lambda\left( \mathcal{L}_{KL}^{1} + \mathcal{L}_{KL}^{2} + \cdots \right)
    \label{equ:L} 
\end{equation}
We believe that the $ \mathcal{L}_{CE} $ to $ \mathcal{L}_{KL} $ weight ratio should not be a constant value and should change with training, so we propose the cosine variation weight of Equation \ref{equ:lam}, which changes as shown in Figure \ref{fig:lam}.
\begin{equation}
    \lambda = \alpha\left( {1 - 0.5\left( {1 + {\cos\left( {\frac{\mathcal{s}}{\mathcal{S}}\pi} \right)}} \right)} \right)
    \label{equ:lam} 
\end{equation}

\subsection{Training Procedure}

The training procedure of SMC is summarized in Algorithm \ref{alg:DLB}. Note that at the beginning of training, there is no soft label saved in the previous step, so it is not possible to calculate $ \mathcal{L}_{KL} $, at this time $ {\mathcal{L} = ~\mathcal{L}}_{CE} $. The data in channel A, channel B, channel C, etc. can be calculated (line \ref{line:4} and line \ref{line:5}) in parallel to get the corresponding soft labels. At each iteration, the loss function is calculated using Equation \ref{equ:LCE}, Equation \ref{equ:LKL} and Equation \ref{equ:L} to compute the gradient information generated by consistency regularization, self-distillation and hard labeling, respectively. Models converging according to this gradient information have better generalization performance and strong robustness to label noise. Note that the SMC uses the same data enhancement method $ \phi $ for data in channel A, channel B, channel C, etc.

However, the augmentation contains random events, which can cause different augmentation results $ \phi\left( x_{A} \right), \phi\left( x_{B} \right), \phi\left( x_{C} \right), $ etc., and this also evaluates the augmentation method several times in time, leading to additional changes in the soft labels.

However, the augmentation contains random events, which can cause the resulting augmentation $ \phi\left( x_{A} \right) $ to be different from $ \phi\left( x_{B} \right) $, and this also evaluates the augmentation method twice in time, resulting in additional variations in the soft labels. By increasing the gap between the soft labels of the same data in channel A, channel B, channel C, etc. in this way, the resulting gradient contains a part of the effect of consistency regularization.

\begin{algorithm}[H]
    \caption{Training with SMC}
    \label{alg:DLB}
    \begin{algorithmic}[1]
    \Require 
    A network $ \mathcal{M} $ with weights $ \omega $; Learing $ \eta $; Epochs $ \mathit{E} $; Iterations $ \mathit{T} $per epoch; SMC coefficients $ \lambda $; SMC hyperparameter $ \alpha $; Temperature $ \tau $; Now step $ \mathfrak{s} $; Total Step $ \mathfrak{S} $
    \For{$ \mathit{e} = 1 $ to $ \mathit{E} $ }{\Comment{ $ \mathit{e} $represents the current epoch}}
        \For{ $ \mathit{t} = 1 $ to $ \mathit{T} $, Sample a mini-batch $ \mathcal{B}_t $ }
        
            \State $ \lambda \gets f\left( {\alpha,\mathfrak{s},\mathfrak{S}} \right) ${\Comment{Defined in Equation \ref{equ:lam} }}
            \State $ \left\{ p_{i}^{1,t} \right\}_{x_{A}} $, $ \left\{ p_{i}^{\tau,t} \right\}_{x_{A}} \gets \mathcal{M}\left( {\phi \left( {x_{A}} \right);\omega_{t}} \right) $ \label{line:4}
            \State $ \left\{ p_{i}^{\tau,t} \right\}_{x_{B}} = \mathcal{M}\left( {\phi\left( x_{B} \right);\omega_{t}} \right),~\left\{ p_{i}^{\tau,t} \right\}_{x_{C}} = \mathcal{M}\left( {\phi\left( x_{C} \right);\omega_{t}} \right),~\cdots $ \label{line:5}
            \State Compute $ \mathcal{L}_{CE} $ {\Comment{Defined in Equation \ref{equ:LCE}}}
            \State Load $ \left\{ p_{i}^{\tau,t - 1} \right\}_{x_{B}},~\left\{ p_{i}^{\tau,t - 1} \right\}_{x_{C}},~\cdots~ $ stored at step $ t - 1 $
            \State Compute $ \mathcal{L}_{KL}^{1},~\mathcal{L}_{KL}^{2},~\cdots ${\Comment{Defined in Equation \ref{equ:LKL}}}
            \State $ \mathcal{L} = \left( {1 - \lambda} \right)\mathcal{L}_{CE} + ~~\lambda\left( \mathcal{L}_{KL}^{1} + \mathcal{L}_{KL}^{2} + \cdots \right) $
            \State Update the weights: $ \omega_{t + 1} \gets \omega_{t} - \eta\nabla_{\omega_{t}}\mathcal{L}|_{\omega_{t}} $
        \EndFor
    \EndFor
\end{algorithmic}
\end{algorithm}

\section{Experiments}

In this section, we verify the effectiveness of SMC. We have divided the SMC into two versions: SMC with channel A and channel B only (SMC-2); SMC with channel A, channel B, and channel C only (SMC-3). We first demonstrated experimentally that SMC-2 has better performance than DLB, LSR and SAM with the same time consumption. Then, we demonstrate that SMC-2 is compatible with better augmentation methods. Next, we demonstrate that SMC-2 can effectively improve the robustness of the model to label noise and provides significant resistance to the reduced generalization ability of the model. Then, we show the effect of hyperparameters on the performance of SMC-2. Finally, we experimentally show that SMC-3 performs slightly better than SMC-2 performance on Cifar100 without adjusting the hyperparameters.

\subsection{Datasets and Settings}

\paragraph{Datasets.}To evaluate the performance of SMC, we used the following commonly used image classification benchmark datasets for experiments : CIFAR-10, CIFAR-100 \cite{krizhevsky2009cifar}, ImageNet \cite{deng2009imagenet}. Cifar10/100 contains 60,000 RGB images of 32$ \times $32 pixels, which are divided into 10/100 classes, each containing 5000/500 training samples and 1000/100 testing samples. ImageNet uses its subset ILSVRC2012, which contains 1.35 million 224$ \times $224 pixel RGB images divided into 1000 categories, each containing about 1000 training samples and 50 testing samples.

\paragraph{Models.}We used VGG19-BN \cite{simonyan2014very}, ResNet-56, ResNet-110 \cite{he2016deep}, WRN28-10 \cite{zagoruyko2016wide}, PreResNet-110 \cite{he2016identity}, DenseNet-100-12 \cite{huang2017densely} for training on the CIFAR-10/100 dataset. We trained on ImageNet dataset using AlexNet \cite{krizhevsky2014one} models.

\paragraph{Baselines.}We use the network with SGD optimizer training hard labels as the baseline, and the specific hyperparameters can be found in the Appendix.

\paragraph{Compared methods.}We compared with Label smoothing regularizaion (LSR) \cite{szegedy2016rethinking}, Sharpness-aware minimization (SAM) \cite{foret2020sharpness} and Self-distillation from the last mini-batch (DLB) \cite{shen2022self}.

\paragraph{Number of channels.}SMC-2 has only channel A and channel B. SMC-3 has only channel A, channel B, and channel C.

\paragraph{Implementations.}When we compare with other methods, the hyperparameters are the same for each model. To ensure fairness, the SMC-2 elapsed time is used as the baseline in the comparison. Among them, although DLB uses twice as much data in calculating $ \mathcal{L}_{CE} $, it only has one more forward propagation time compared with SGD in terms of time consumption, which is the same as SMC-2 time consumption, and therefore does not cut its epochs. If $ \mathcal{C}_A $ and $ \mathcal{C}_B $ data are not parallelized when using SMC-2 to get soft labels through the model, then SAM consumes one more time for backward propagation compared with SMC-2, then we consider the worst case to set its epochs to 150. Of course, considering the worst case, the epochs of LSR are set to 300. We adopt the standard data augmentation scheme for Cifar-10/100: 32$ \times $32 random crop after padding with 4 pixels; random horizontal flip; normalized by deviation. For ImageNet, we adopt the standard data augmentation scheme: 224$ \times $224 random crop; random horizontal flip; normalized by deviation. For the datasets CIFAR-100, we set the hyperparameters to $ \tau = 1.5 $ and $ \alpha = 0.9 $. For the dataset CIFAR-10 and ImageNet, we set the hyperparameters to $ \tau = 1.0 $ and $ \alpha = 0.9 $. Other specific hyperparameters can be found in Appendix. 

\begin{table}[htbp]
    \caption{Comparison of the Top-1 Accuracy (\%) of SMC-2 with other methods on the CIFAR-100 dataset. The best performance is highlighted in boldface. We calculated the mean and deviation by running three different seeds.}
    \label{tab:c100}
    \centering
    \begin{tabularx}{\textwidth}{ZZZZZZZ}
    \toprule
    \multicolumn{1}{c}{\multirow{3.5}{*}{Methods}} &
    \multicolumn{6}{c}{Models}  \\
    \cmidrule(l){2-7}
      & VGG19-BN & ResNet-56 & ResNet-110 & PreResNet-110 & WRN-28-10 & DenseNet-100-12 \\
    \midrule
    Vanilla & $ {73.65}_{\pm 0.10} $ & $ {70.89}_{\pm 0.27} $ & $ {72.02}_{\pm 0.15} $ & $ {72.85}_{\pm 0.24} $ & $ {81.40}_{\pm 0.07} $ & $ {76.71}_{\pm 0.06} $ \\
    LSR & $ {74.53}_{\pm 0.19} $ & $ {71.49}_{\pm 0.09} $ & $ {72.31}_{\pm 0.97} $ & $ {73.92}_{\pm 0.26} $ & $ {80.63}_{\pm 0.04} $ & $ {76.85}_{\pm 0.10} $ \\ 
    DLB & $ {\textbf{75.09}}_{\pm 0.42} $ & $ {70.02}_{\pm 0.29} $ & $ {71.65}_{\pm 0.23} $ & $ {72.49}_{\pm 0.24} $ & $ {81.24}_{\pm 0.36} $ & $ {76.08}_{\pm 0.25} $ \\
    SAM & $ {73.24}_{\pm 0.20} $ & $ {70.97}_{\pm 0.38} $ & $ {72.59}_{\pm 0.50} $ & $ {73.81}_{\pm 0.30} $ & $ {\textbf{82.71}}_{\pm 0.13} $ & $ {77.47}_{\pm 0.31} $ \\
    \midrule
    SMC-2 & $ {\textbf{75.09}}_{\pm 0.19} $ & $ {\textbf{72.24}}_{\pm 0.10} $ & $ {\textbf{73.09}}_{\pm 1.04} $ & $ {\textbf{74.72}}_{\pm 0.20} $ & $ {82.14}_{\pm 0.19} $ & $ {\textbf{78.10}}_{\pm 0.14} $\\
    (Ours) & $ \left( 1.44 \uparrow \right) $ & $ \left( 1.35\uparrow \right) $ & $ \left( 1.07\uparrow \right) $ & $ \left( 1.87\uparrow \right) $ & $ \left( 0.74\uparrow \right) $ & $ \left( 1.39\uparrow \right) $ \\ 
    \bottomrule
    \end{tabularx}
\end{table}

\subsection{Experimental Results}

From Table \ref{tab:c100}, it can be seen that the improvement of SMC-2 on Cifar100 relative to vanilla its validation top-1 accuracy ranges from 0.74\% to 1.87\%, and on Cifar10 compared to vanilla its validation top-1 accuracy ranges from 0.26\% to 0.65\%, which shows that SMC-2 is effective. SMC-2 has better performance results on Cifar10/100 when compared to other competitors (LSR, DLB)\footnote{We do not consider SAM as our competitor, because in principle we are compatible with SAM. We introduced SAM because SMC-2 has a certain effect of reducing loss of sharpness and to demonstrate the performance effect of SMC-2.}. Also compared to the advanced method SAM, our algorithm converges faster and better except for the WRN-28-10 model.

\begin{figure}[htbp]
    \centering
    \includegraphics[height = 4cm, width = 6cm]{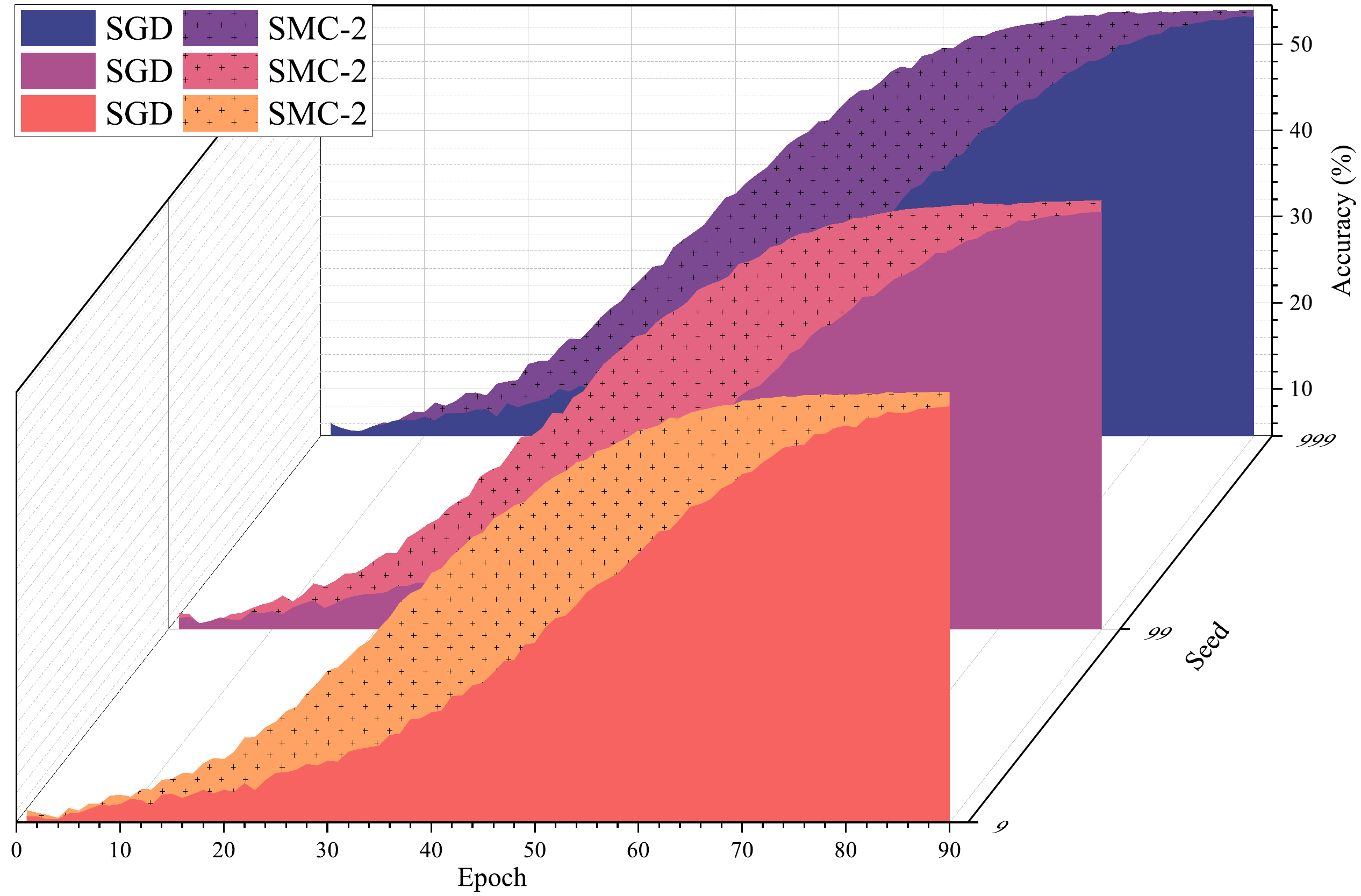}
    \caption{Comparison of the validation Top-1 accuracy results of SGD and SMC-2 trained on ImageNet with AlexNet model using different seeds. In the figure, SGD is without filling pattern and SMC-2 is with filling pattern. SGD and SMC-2 with the same seed will be grouped together. SGD is in the front of the group and SMC-2 is in the back.}
    \label{fig:ImageNet}
\end{figure}

As can be seen in Figure \ref{fig:ImageNet}, SMC-2 is still effective on large data sets. Compared with SGD, SMC-2 significantly improves the convergence speed and generalization ability. However, due to the limitation of our computational resources, we cannot choose the hyperparameters of SMC-2 suitable for ImageNet, so we cannot compare it with other methods.

\begin{table}[htbp]
    \caption{Comparison of the Top-1 Accuracy (\%) of SMC-2 with other methods on the CIFAR-10 dataset. The best performance is highlighted in boldface. We calculated the mean and deviation by running three different seeds.}
    \label{tab:c10}
    \centering
    \begin{tabularx}{\textwidth}{ZZZZZZZ}
    \toprule
    \multicolumn{1}{c}{\multirow{3.5}{*}{Methods}} &
    \multicolumn{6}{c}{Models}  \\
    \cmidrule(l){2-7}
      & VGG19-BN & ResNet-56 & ResNet-110 & PreResNet-110 & WRN-28-10 & DenseNet-100-12 \\
    \midrule
    Vanilla & $ {93.73}_{\pm 0.38} $ & $ {93.57}_{\pm 0.21} $ & $ {93.96}_{\pm 0.19} $ & $ {94.37}_{\pm 0.09} $ & $ {96.30}_{\pm 0.05} $ & $ {95.18}_{\pm 0.15} $ \\
    LSR & $ {94.04}_{\pm 0.16} $ & $ {93.95}_{\pm 0.11} $ & $ {\textbf{94.34}}_{\pm 0.32} $ & $ {94.58}_{\pm 0.10} $ & $ {96.19}_{\pm 0.12} $ & $ {95.20}_{\pm 0.15} $ \\ 
    DLB & $ {94.05}_{\pm 0.03} $ & $ {93.36}_{\pm 0.28} $ & $ {93.05}_{\pm 1.06} $ & $ {94.21}_{\pm 0.06} $ & $ {96.38}_{\pm 0.12} $ & $ {95.53}_{\pm 0.13} $ \\
    SAM & $ {94.37}_{\pm 0.29} $ & $ {93.68}_{\pm 0.20} $ & $ {94.29}_{\pm 0.14} $ & $ {94.65}_{\pm 0.06} $ & $ {\textbf{96.78}}_{\pm 0.09} $ & $ {95.48}_{\pm 0.19} $ \\
    \midrule
    SMC-2 & $ {\textbf{94.38}}_{\pm 0.09} $ & $ {\textbf{94.08}}_{\pm 0.16} $ & $ {\textbf{94.34}}_{\pm 0.22} $ & $ {\textbf{94.71}}_{\pm 0.18} $ & $ {96.56}_{\pm 0.03} $ & $ {\textbf{95.56}}_{\pm 0.14} $\\
    (Ours) & $ \left( 0.65 \uparrow \right) $ & $ \left( 0.51\uparrow \right) $ & $ \left( 0.39\uparrow \right) $ & $ \left( 0.34\uparrow \right) $ & $ \left( 0.26\uparrow \right) $ & $ \left( 0.38\uparrow \right) $ \\ 
    \bottomrule
    \end{tabularx}
\end{table}    

\subsubsection{Symbiosis with Augmentations}

SMC-2 evaluates the augmentation twice at the same time, so it will be more effective for some Augmentations . We use AutoAugment (AA) \cite{cubuk2018autoaugment}, CutOut \cite{devries2017improved} to enhance the dataset Cifar100 to evaluate the symbiotic effect of SMC-2 on the mutual achievement of data augmentation. From Table 4 we can observe that the SMC-2 algorithm has an average augmentation effect of 1.27\% and 0.87\% on AA and Cutout, so SMC-2 also has a good performance improvement for some of the augmentations.

\begin{table}[htbp]
    \caption{Symbiotic effect of SMC-2 with different augmentations on the CIFAR-100 dataset. The best performance is highlighted in boldface. We calculated the mean and deviation by running three different seeds.}
    \label{tab:aug}
    \centering
    \begin{tabularx}{\textwidth}{ZZZZZZZ}
    \toprule
    \multicolumn{1}{c}{\multirow{3.5}{*}{Methods}} &
    \multicolumn{6}{c}{Models}  \\
    \cmidrule(l){2-7}
      & VGG19-BN & ResNet-56 & ResNet-110 & PreResNet-110 & WRN-28-10 & DenseNet-100-12 \\
    \midrule
    Vanilla & $ {73.65}_{\pm 0.10} $ & $ {70.89}_{\pm 0.27} $ & $ {72.02}_{\pm 0.15} $ & $ {72.85}_{\pm 0.24} $ & $ {81.40}_{\pm 0.07} $ & $ {76.71}_{\pm 0.06} $ \\
    \multirow{2}{*}{+SMC-2} & $ {\textbf{75.09}}_{\pm 0.19} $ & $ {\textbf{72.24}}_{\pm 0.10} $ & $ {\textbf{73.09}}_{\pm 1.04} $ & $ {\textbf{74.72}}_{\pm 0.20} $ & $ {\textbf{82.14}}_{\pm 0.19} $ & $ {\textbf{78.10}}_{\pm 0.14} $ \\
    & $ \left( 1.44 \uparrow \right) $ & $ \left( 1.35\uparrow \right) $ & $ \left( 1.07\uparrow \right) $ & $ \left( 1.87\uparrow \right) $ & $ \left( 0.74\uparrow \right) $ & $ \left( 1.39\uparrow \right) $ \\ 
    \midrule
    CutOut & $ {74.60}_{\pm 0.48} $ & $ {71.40}_{\pm 0.26} $ & $ {72.94}_{\pm 0.76} $ & $ {74.13}_{\pm 0.15} $ & $ {82.15}_{\pm 0.15} $ & $ {77.97}_{\pm 0.18} $ \\
    \multirow{2}{*}{+SMC-2} & $ {\textbf{75.97}}_{\pm 0.21} $ & $ {\textbf{73.20}}_{\pm 0.19} $ & $ {\textbf{74.32}}_{\pm 0.43} $ & $ {\textbf{75.58}}_{\pm 0.27} $ & $ {\textbf{83.17}}_{\pm 0.17} $ & $ {\textbf{78.62}}_{\pm 0.32} $ \\
    & $ \left( 1.37 \uparrow \right) $ & $ \left( 1.80\uparrow \right) $ & $ \left( 1.38\uparrow \right) $ & $ \left( 1.45\uparrow \right) $ & $ \left( 1.02\uparrow \right) $ & $ \left( 0.64\uparrow \right) $ \\ 
    \midrule 
    AA & $ {75.65}_{\pm 0.44} $ & $ {74.00}_{\pm 0.45} $ & $ {75.54}_{\pm 0.37} $ & $ {76.11}_{\pm 0.35} $ & $ {82.79}_{\pm 0.23} $ & $ {79.76}_{\pm 0.12} $ \\
    \multirow{2}{*}{+SMC-2} & $ {\textbf{76.56}}_{\pm 0.25} $ & $ {\textbf{74.83}}_{\pm 0.29} $ & $ {\textbf{76.64}}_{\pm 0.37} $ & $ {\textbf{76.86}}_{\pm 0.24} $ & $ {\textbf{84.14}}_{\pm 0.03} $ & $ {\textbf{80.04}}_{\pm 0.12} $ \\
    & $ \left( 0.92 \uparrow \right) $ & $ \left( 0.82\uparrow \right) $ & $ \left( 1.10\uparrow \right) $ & $ \left( 0.76\uparrow \right) $ & $ \left( 1.35\uparrow \right) $ & $ \left( 0.28\uparrow \right) $ \\ 
    \bottomrule
    \end{tabularx}
\end{table} 

\subsubsection{Robustness to Data Corruption}

SMC-2 has access to historical information and has consistent regularization properties, so it can significantly increase the robustness of neural networks to label noise. To verify this idea, we use VGG19-BN, ResNet-56, WRN28-10, PreResNet-110 models trained on Cifar-10/100 with added label noise. We injected different proportions $ \eta = \left\{ 10\%,~20\%,~30\%,~40\% \right\} $ of label noise in the training set respectively in a manner consistent with previous work \cite{arazo2019unsupervised}.

\begin{figure}[htbp]
    \centering
    \begin{subfigure}{0.2\linewidth}
        \centering
        \includegraphics[width=0.9\linewidth]{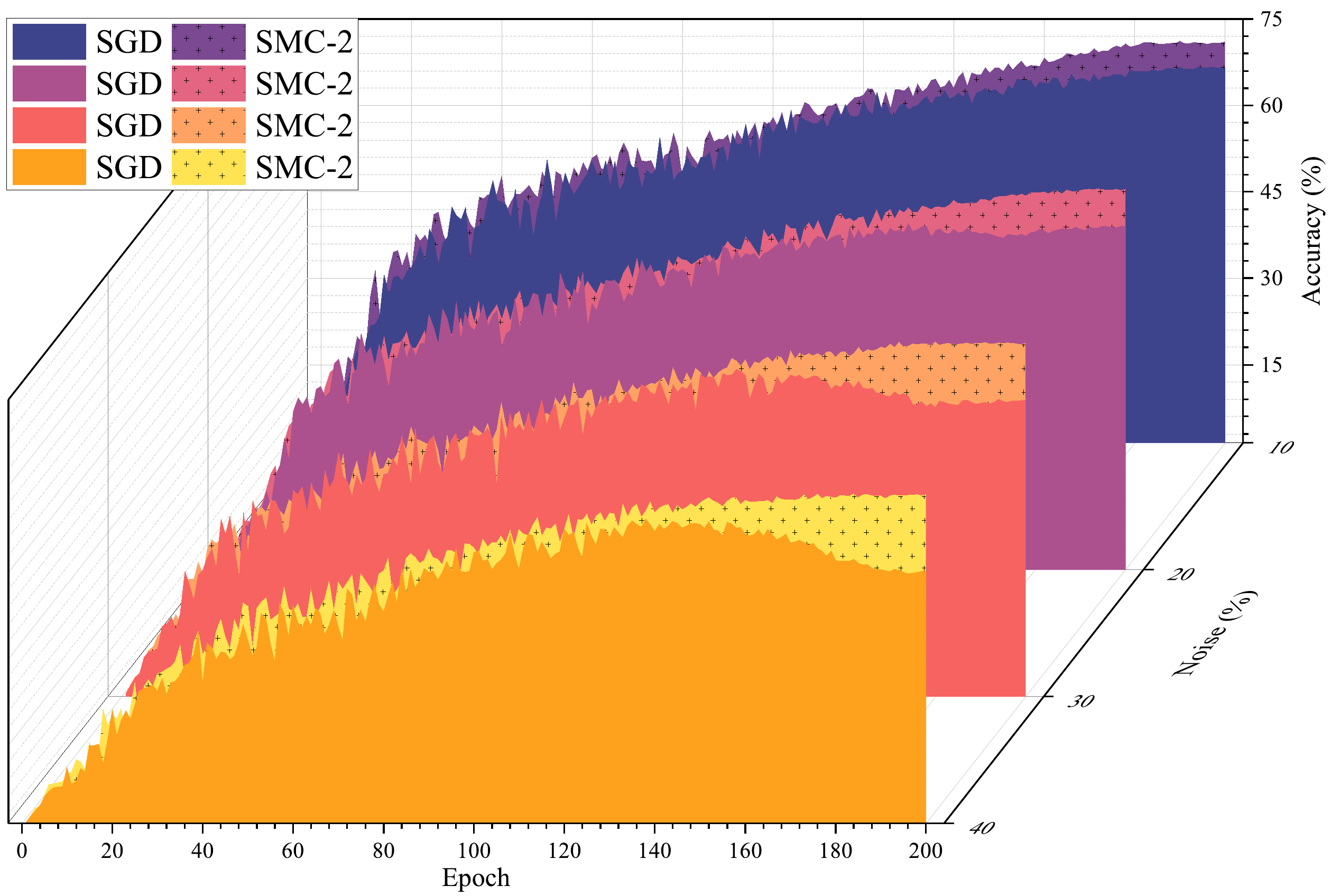}
        \caption{VGG19-BN}
        \label{fig:noise-c100-vgg}
    \end{subfigure}
    \centering
    \begin{subfigure}{0.2\linewidth}
        \centering
        \includegraphics[width=0.9\linewidth]{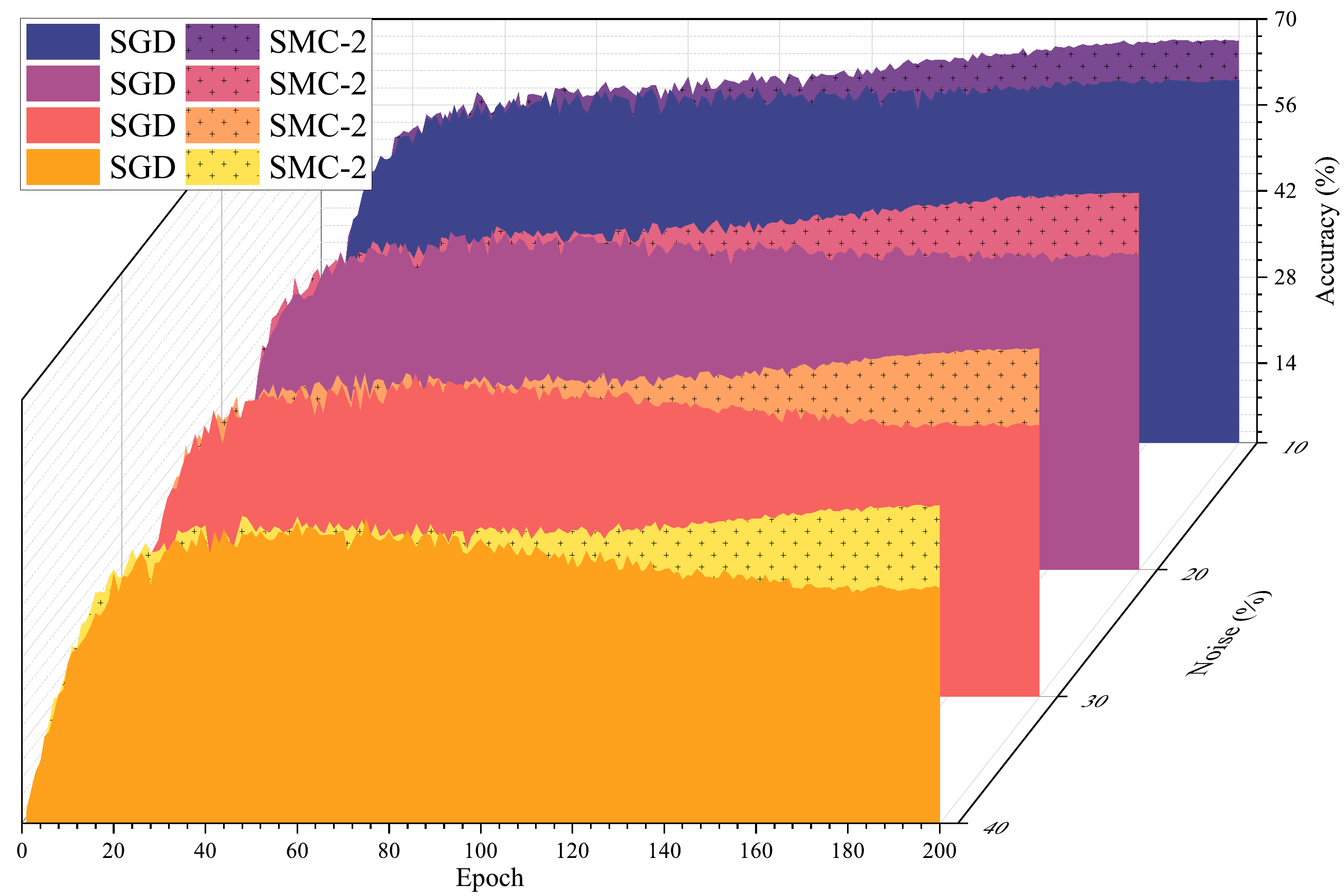}
        \caption{ResNet-56}
        \label{fig:noise-c100-R56}
    \end{subfigure}
    \centering
    \begin{subfigure}{0.2\linewidth}
        \centering
        \includegraphics[width=0.9\linewidth]{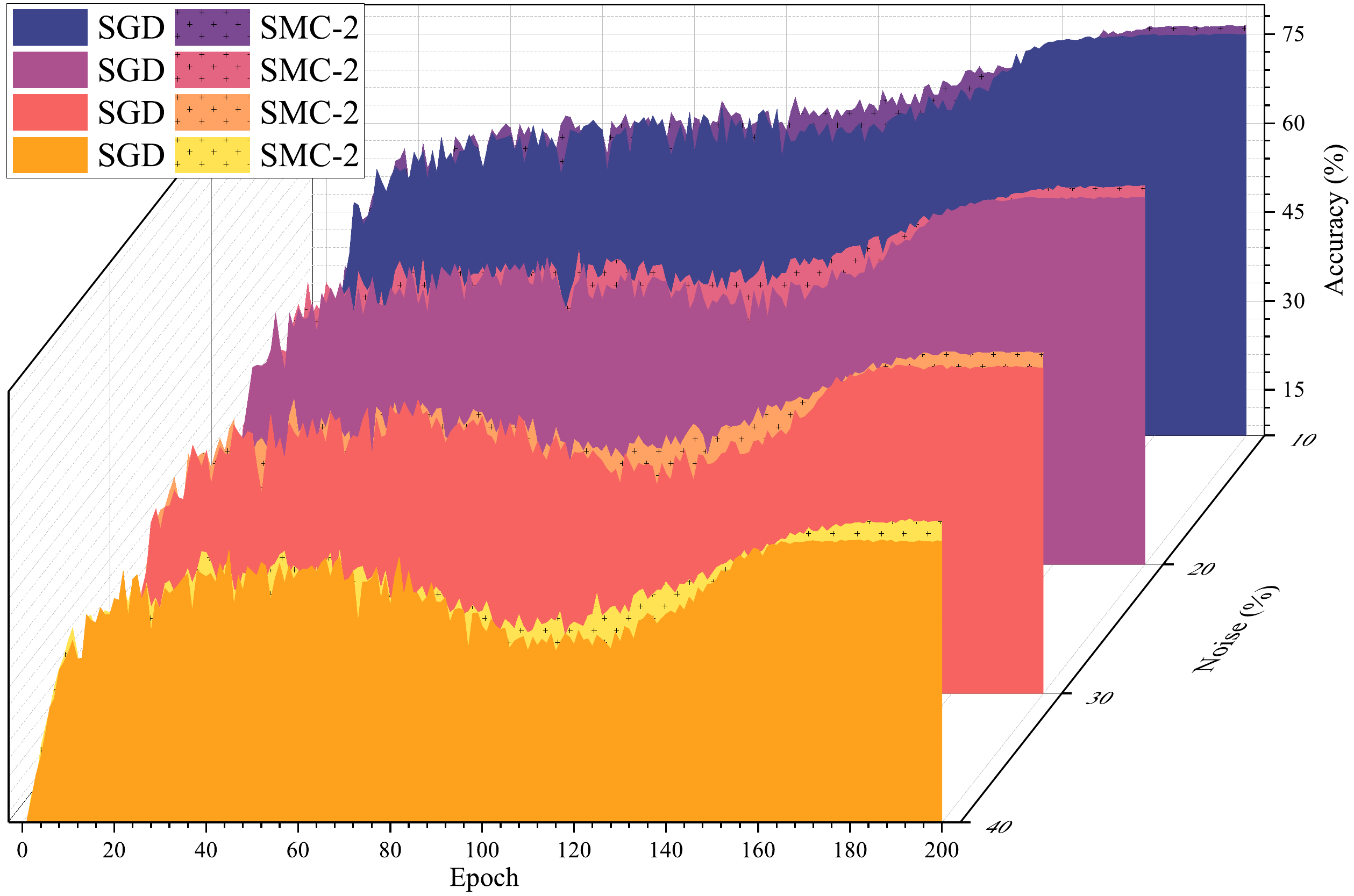}
        \caption{WRN-28-10}
        \label{fig:noise-c100-WRN}
    \end{subfigure}
    \centering
    \begin{subfigure}{0.2\linewidth}
        \centering
        \includegraphics[width=0.9\linewidth]{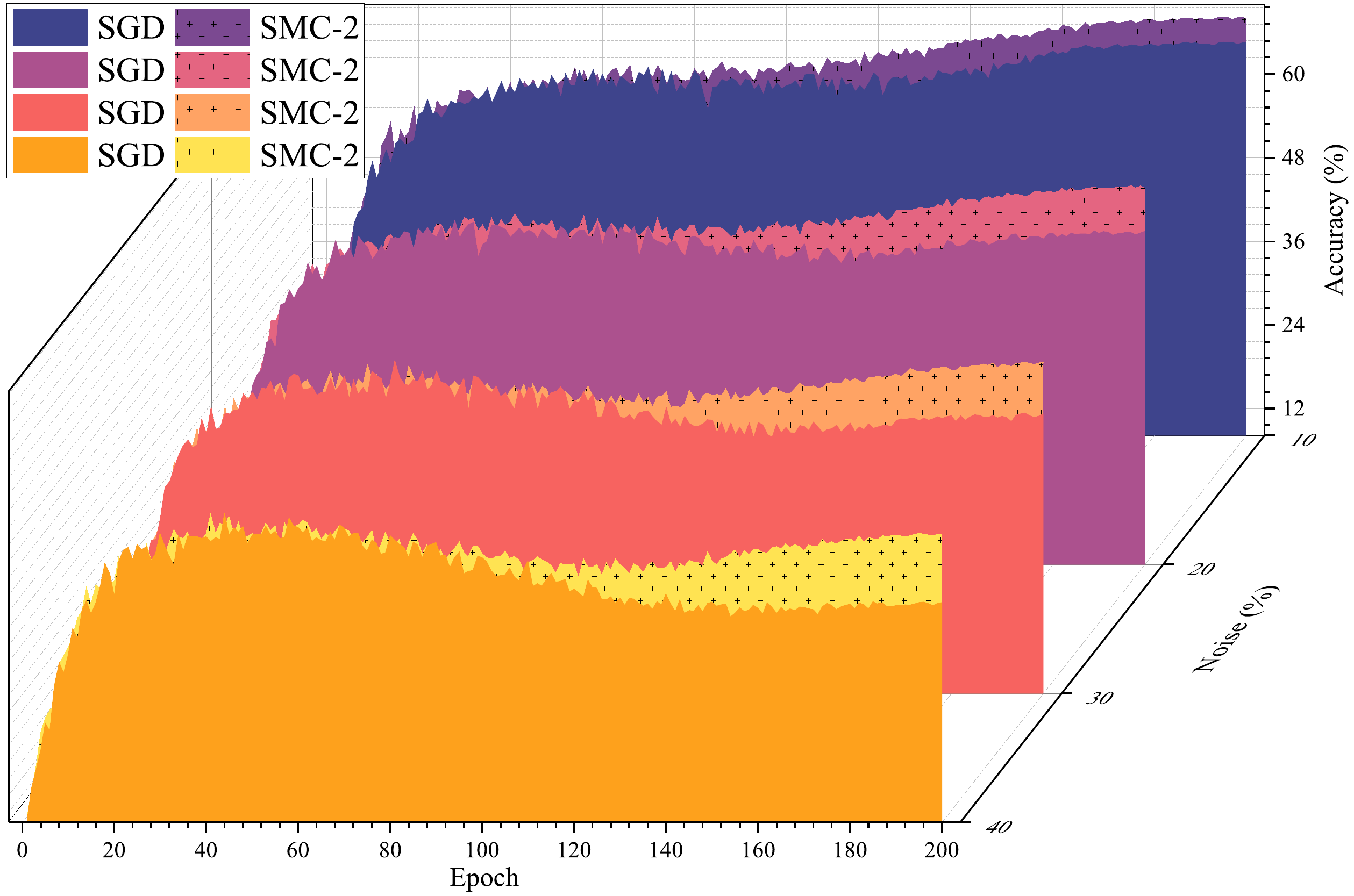}
        \caption{PreResNet-110}
        \label{fig:noise-c100-PreRes}
    \end{subfigure}
    \caption{SMC-2 improves resistance to label noise when the model is trained on CIFAR-100 with different intensities $ \eta $ of label noise. In the figure, SGD is without filling pattern and SMC-2 is with filling pattern. SGD and SMC-2 with the same label noise level $ \eta $ will be grouped together. SGD is in the front of the group and SMC-2 is in the back.}
    \label{fig:noise-c100}
\end{figure}

\begin{figure}[htbp]
    \centering
    \begin{subfigure}{0.2\linewidth}
        \centering
        \includegraphics[width=0.9\linewidth]{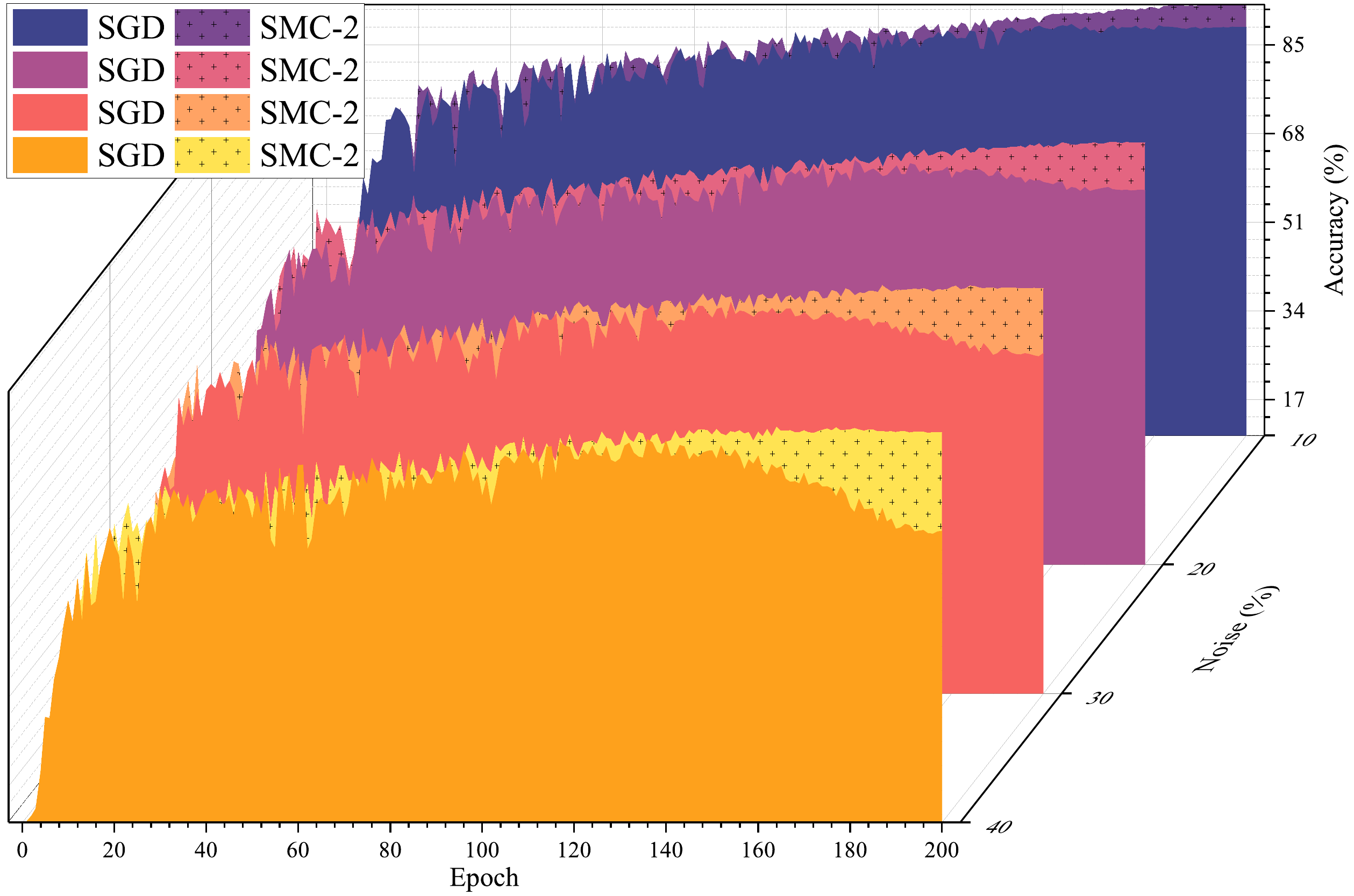}
        \caption{VGG19-BN}
        \label{fig:noise-c10-vgg}
    \end{subfigure}
    \centering
    \begin{subfigure}{0.2\linewidth}
        \centering
        \includegraphics[width=0.9\linewidth]{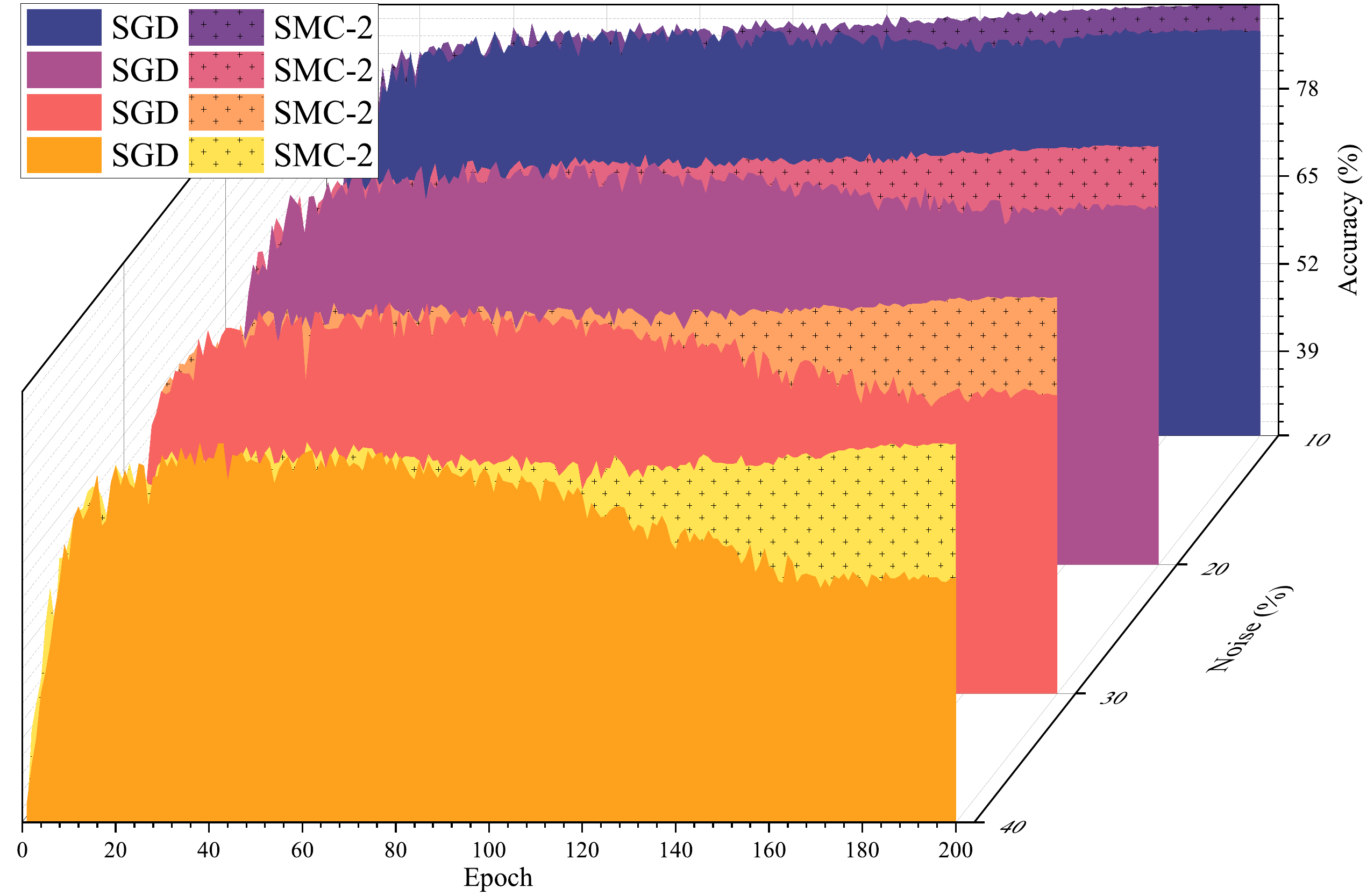}
        \caption{ResNet-56}
        \label{fig:noise-c10-R56}
    \end{subfigure}
    \centering
    \begin{subfigure}{0.2\linewidth}
        \centering
        \includegraphics[width=0.9\linewidth]{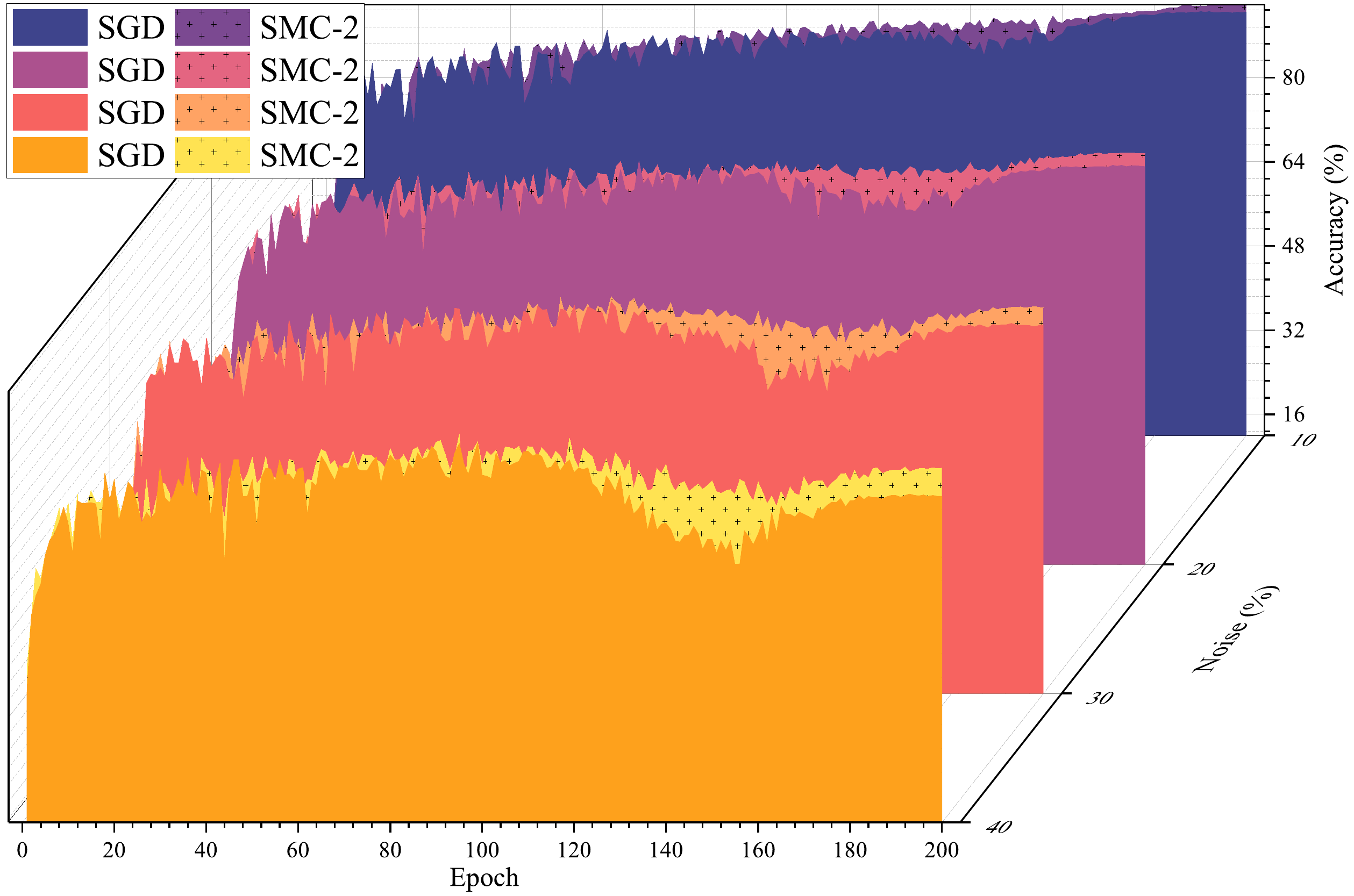}
        \caption{WRN-28-10}
        \label{fig:noise-c10-WRN}
    \end{subfigure}
    \centering
    \begin{subfigure}{0.2\linewidth}
        \centering
        \includegraphics[width=0.9\linewidth]{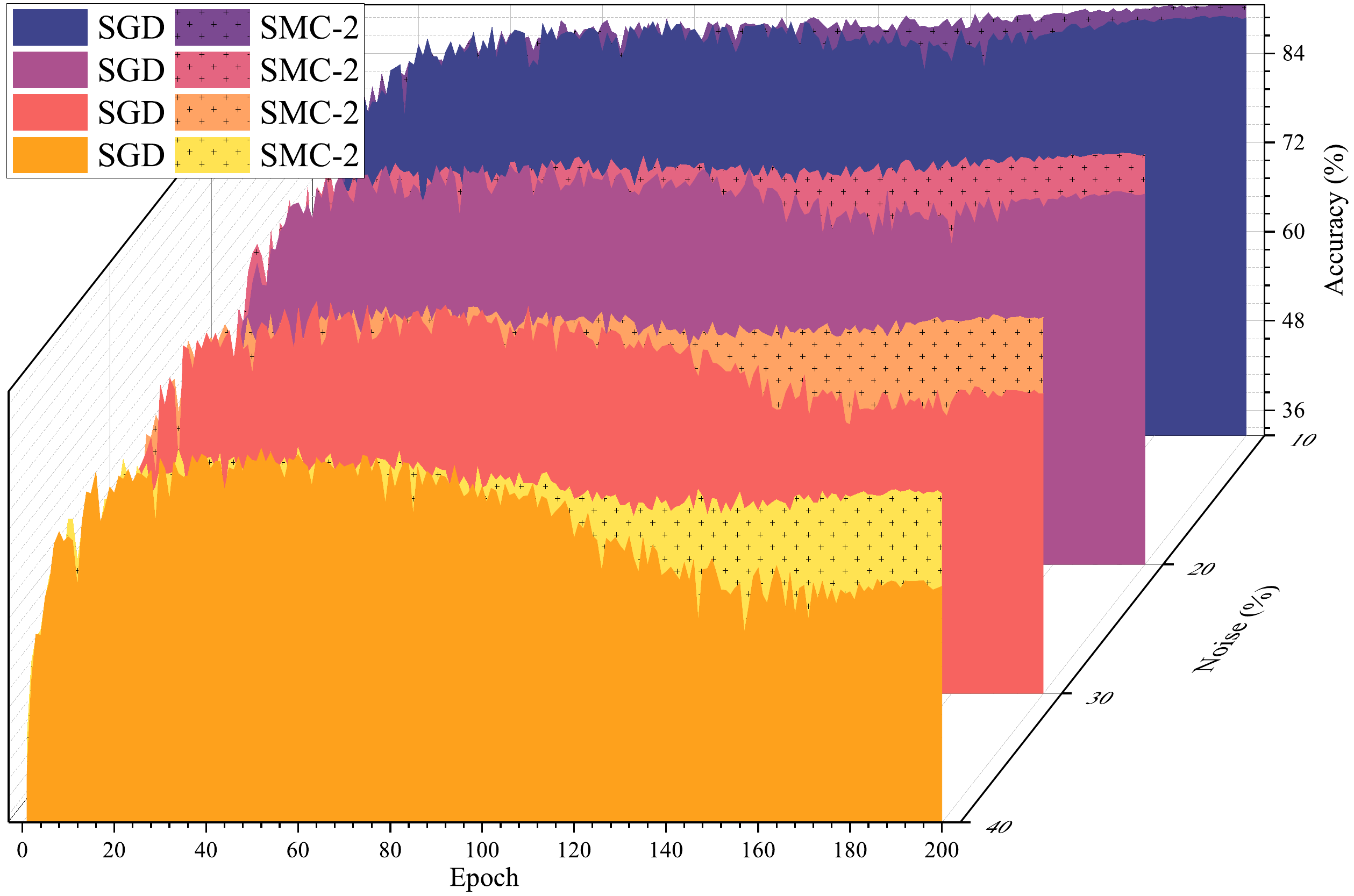}
        \caption{PreResNet-110}
        \label{fig:noise-c10-PreRes}
    \end{subfigure}
    \caption{SMC-2 improves resistance to label noise when the model is trained on CIFAR-10 with different intensities $ \eta $ of label noise. In the figure, SGD is without filling pattern and SMC-2 is with filling pattern. SGD and SMC-2 with the same label noise level $ \eta $ will be grouped together. SGD is in the front of the group and SMC-2 is in the back.}
    \label{fig:noise-c10}
\end{figure}

From Figures \ref{fig:noise-c100} and \ref{fig:noise-c10}, we can observe that SMC-2 does not increase and then decrease the accuracy during training for models with poor noise robustness, and can significantly improve the best accuracy. For example, for the VGG19-BN network model, we improved its optimal accuracy by \{10.89\%, 11.41\%, 9.51\%, 8.08\%\} at $ \eta = \left\{ 10\%,~20\%,~30\%,~40\% \right\} $, respectively.

\subsection{Analysis about Hyper-Parameters}

\subsubsection{Analysis of  ~\texorpdfstring{$ \alpha $}{alpha}}

We first investigate the effect of $ \alpha $ on the performance of SMC-2 for $ \tau =5 $. Training the WRN-28-10 model on the CIFAR-100 dataset, using SGD as the optimizer, we show the validation top-1 accuracy for different $ \alpha(0.1,~0.2,~0.3,\cdots,~0.8,~0.9) $ in Figure \ref{fig:alpha}. It can be seen that the accuracy reaches 81.7\% for $ \alpha =0.9 $. Of course, when $ \alpha \in \lbrack 0.8,~1.0\rbrack $, SMC-2 has good performance in all cases where the training set accuracy is high ( the hyperparameter should be reduced when the training set accuracy is not high).

\subsubsection{Analysis of ~\texorpdfstring{$ \tau $}{tau}}

\begin{figure}[htbp]
    \centering
    \begin{subfigure}{0.4\linewidth}
        \centering
        \includegraphics[width=0.595\linewidth]{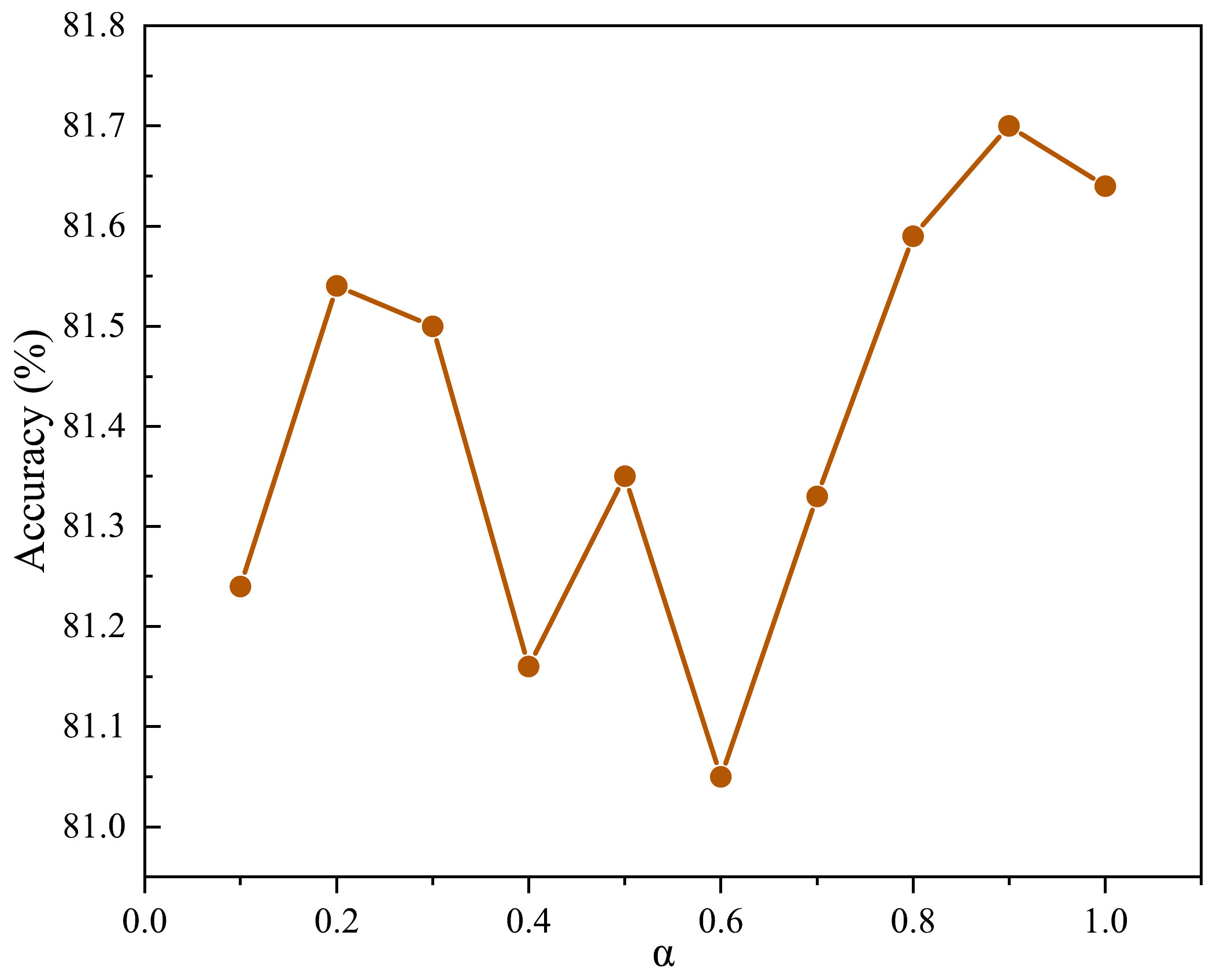}
        \caption{Effect of $ \alpha $}
        \label{fig:alpha}
    \end{subfigure}
    \centering
    \begin{subfigure}{0.4\linewidth}
        \centering
        \includegraphics[width=0.63\linewidth]{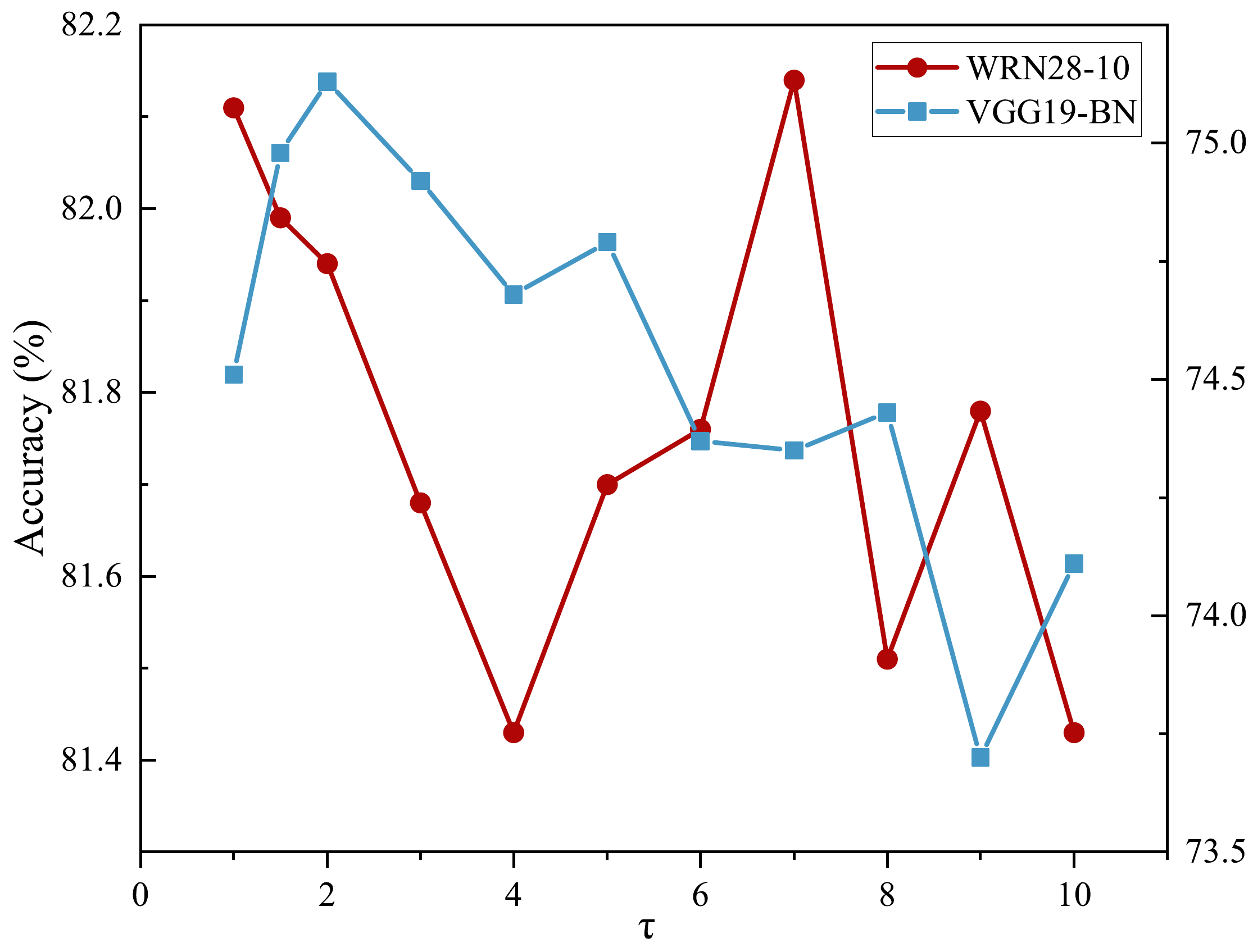}
        \caption{Effect of $ \tau $}
        \label{fig:tau}
    \end{subfigure}
    \caption{Effect of hyperparameters for SMC-2 performance on the CIFAR-100 dataset.}
    \label{fig:hyper}
\end{figure}

Then, we make $ \alpha =0.9 $ and train the WRN28-10 and VGG19-BN models on the dataset Cifar-100, whose generalization ability changes when $ \tau(1,2,3,\cdots,10) $ is varied as in Figure \ref{fig:tau}. We observe that SMC-2 performs well on both models for $ \tau \in \lbrack 1,2\rbrack $. However, to examine the performance of SMC-2, we chose $ \tau = 1.5 $  as the hyperparameter of SMC-2 on the dataset Cifar-100 and $ \tau = 1.0 $  as the hyperparameter of SMC-2 on the dataset Cifar-10.

\subsection{ Analysis of number of channels}

We make the hyperparameters of SMC-2 and SMC-3 the same, and the comparison of their performance on Cifar-100 is shown in Table \ref{tab:channel}. We can observe that SMC-3 makes the generalization ability of most models better than SMC-2. However, in order to compare the performance with other methods, we choose SMC-2, which takes less time and still has better performance, to compare with other methods.

\begin{table}[htbp]
    \caption{The effect of the number of channels for the SMC performance on the dataset Cifar-100. The best performance is highlighted in boldface. We calculated the mean and deviation by running three different seeds.}
    \label{tab:channel}
    \centering
    \begin{tabularx}{\textwidth}{ZZZZZZZ}
    \toprule
    \multicolumn{1}{c}{\multirow{3.5}{*}{Methods}} &
    \multicolumn{6}{c}{Models}  \\
    \cmidrule(l){2-7}
      & VGG19-BN & ResNet-56 & ResNet-110 & PreResNet-110 & WRN-28-10 & DenseNet-100-12 \\
    \midrule
    Vanilla & $ {73.65}_{\pm 0.10} $ & $ {70.89}_{\pm 0.27} $ & $ {72.02}_{\pm 0.15} $ & $ {72.85}_{\pm 0.24} $ & $ {81.40}_{\pm 0.07} $ & $ {76.71}_{\pm 0.06} $ \\
    \midrule
    \multirow{2}{*}{SMC-2} & $ {75.09}_{\pm 0.19} $ & $ {72.24}_{\pm 0.10} $ & $ {73.09}_{\pm 1.04} $ & $ {\textbf{74.72}}_{\pm 0.20} $ & $ {82.14}_{\pm 0.19} $ & $ {\textbf{78.10}}_{\pm 0.14} $ \\
    & $ \left( 1.44 \uparrow \right) $ & $ \left( 1.35\uparrow \right) $ & $ \left( 1.07\uparrow \right) $ & $ \left( 1.87\uparrow \right) $ & $ \left( 0.74\uparrow \right) $ & $ \left( 1.39\uparrow \right) $ \\ 
    \midrule
    \multirow{2}{*}{SMC-3} & $ {\textbf{75.69}}_{\pm 0.12} $ & $ {\textbf{72.85}}_{\pm 0.22} $ & $ {\textbf{73.75}}_{\pm 0.31} $ & $ {74.63}_{\pm 0.03} $ & $ {\textbf{82.34}}_{\pm 0.06} $ & $ {77.58}_{\pm 0.24} $ \\
    & $ \left( 2.04 \uparrow \right) $ & $ \left( 1.96\uparrow \right) $ & $ \left( 1.73\uparrow \right) $ & $ \left( 1.79\uparrow \right) $ & $ \left( 0.94\uparrow \right) $ & $ \left( 0.87\uparrow \right) $ \\ 
    \bottomrule
    \end{tabularx}
\end{table} 

\section{Conclusions}

In this work, we propose self-discipline on multiple channels (SMC). The multi-channel concept in SMC enables the simultaneous application of self-distillation and consistency regularization.SMC-2 converges faster and improves model generalization performance more than the state-of-the-art SAM on two datasets. More importantly, SMC-2 significantly improves the robustness of the neural network model to label noise. The introduction of SMC-2 stifled the tendency of the model's generalization ability to increase and then decrease under the influence of label noise. Due to the limitation of computational resources, we could not find suitable hyperparameters $ \tau $ and $ \alpha $ on ImageNet, so we could not compare the performance effect of SMC-2 with other methods. Meanwhile, SMC's formula for calculating the total loss function $ \mathcal{L} $ is somewhat less than perfect. In the future, we will further improve these two aspects. In addition, we will explore the possibility of combining SMC with SAM and its variants.

{\small
\bibliographystyle{ieee_fullname}
\bibliography{egbib.bib}
}

\end{document}